# A Corrective Training Algorithm for Adaptive Learning in Bag Generation

**Hsin-Hsi Chen**     and     **Yue-Shi Lee**

Department of Computer Science and Information Engineering
National Taiwan University
Taipei, Taiwan, R.O.C.
E-mail: hh_chen@csie.ntu.edu.tw

## Abstract

The sampling problem in training corpus is one of the major sources of errors in corpus-based applications. This paper proposes a corrective training algorithm to best-fit the run-time context domain in the application of bag generation. It shows which objects to be adjusted and how to adjust their probabilities. The resulting techniques are greatly simplified and the experimental results demonstrate the promising effects of the training algorithm from generic domain to specific domain. In general, these techniques can be easily extended to various language models and corpus-based applications.

## 1. Introduction

In corpus-based applications, most of the errors are caused by two major sources. One is the power of language models, and the other one is the sampling problem in training corpus. One of the possible ways to avoid the former type of errors is to enhance the weaker language models. The latter type of errors results from the small corpus size and the variant run-time context domain. Small corpus will produce zero and unreliable probabilities in the training tables. Some *smoothing* techniques (Jelinek and Mercer, 1980; Jelinek, 1985; Katz, 1987) have been proposed to deal with this problem. They provide static adjustments of unreliable probabilities. Nevertheless, these methods cannot handle the run-time status of the context domain. *Dynamic* models such as *cache-based* model (Kuhn and Mori, 1990) and *multiple language* model (Matsunaga, *et al.*, 1992) touch on run-time behavior. Cache-based model reflects short-term patterns of words, so that it is effective for repeated expressions. However, this approach is still very intuitive and simple because it only adjusts the word frequencies in run-time, and does not revise the statistical information in the long-term memory. Multiple language model is based on several corpora of different fields. Basically, a small amount of similar text are imported and interpolated with the original texts, when the context domain is presented. The extra cost of this approach is the context determination. The similarity measures among test sentences and the pre-defined context domains may introduce additional errors.

In this paper, we would not like to touch on the power of language models. We focus on the sampling problem in training corpus. A corrective training algorithm, which can be also regarded as a dynamic adaptive learning algorithm, is proposed for bag generation. It exploits the run-time feedback information to best-fit the run-time environment. That is, when error occurs, the error result will be corrected by users. Through the modification, the system learns and adapts. It learns the differences between the correct result and the error result. These form the useful run-time feedback information. In other words, the system learns from the mistakes it makes. Under this way, we first propose a language model to deal with the sentence generation, i.e., *bag generation*, problem and a generic corpus is used to extract the corresponding statistics information. Then the training algorithm will try to adapt the *generic language model* into a specific one according to the useful run-time feedback information. At the same time, the probabilities of the related entries in training table are adjusted. In the following sections we first introduce the bag generation algorithm, then describe the adaptive learning model for bag generation. Before concluding we demonstrate the experimental results of this corrective training algorithm.

## 2. Bag Generation Algorithm

Bag generation (Brown, *et al*., 1990; Chen and Lee, 1993a) is a natural language generation method. It can be applied to develop a generator in a statistically based machine-translation system (Brown, *et al*., 1990; Chen and Lee, 1993b). In bag generation we take a sentence, divide it into words, place the words in a *bag*, and then try to recover the sentence given the bag. That is, given a bag of n words, it tries to find a permutation ρ such that the word sequence <*,$w_{\rho(1)}$, $w_{\rho(2)}$, ...,$w_{\rho(n)}$,*> denotes the correct sentence. The symbol * marks the beginning ($w_{\rho(0)}$) and ending ($w_{\rho(n+1)}$) of a sentence. In Markov word *m*-gram model, the permutation ρ is defined by the following formula.

$$\rho = \arg\max_{\rho} P(w_{\rho(0)}) * P(w_{\rho(1)}|w_{\rho(0)}) * ... * P(w_{\rho(m-2)}|w_{\rho(0)}, ..., w_{\rho(m-3)}) * \prod_{i=m-2}^{n} P(w_{\rho(i+1)}|w_{\rho(i-m+2)}, ..., w_{\rho(i)})$$

Intuitively, a bag generation algorithm can first generate all permutations of words, and then select the permutation with the greatest probability. However, the computational time cannot be endured[1] when the number of words in the bag is large. Here, a *dynamic programming* technique is adopted. For any two sequences,

Sequence 1: *,$w_{\rho'(1)}$, $w_{\rho'(2)}$, ...,$w_{\rho'(i)}$    $1 \leq i \leq n$

Sequence 2: *,$w_{\rho''(1)}$, $w_{\rho''(2)}$, ...,$w_{\rho''(j)}$    $1 \leq j \leq n$

the *merge* operation can be applied in these two sequences under the following four conditions if a Markov word *m*-gram model is used.

(1) The sequence length should be longer than *m*-1, i.e., $i>m$-1 and $j>m$-1.
(2) The lengths of these two sequences should be equal, i.e., $i=j$.
(3) The last *m*-1 words in these two sequences should be equal, i.e.,
    $w_{\rho'(k)}=w_{\rho''(k)}$ for $i$-($m$-1)+1≤$k$≤$i$.
(4) These two sequences should cover the same words, i.e., $w_{\rho'(x)}=w_{\rho''(y)}$ for $1\leq x,y \leq i$.

The merge operation retains the sequence with greater probability, and discards the sequence with smaller probability. The following proposition that clarifies this point for a Markov word *m*-gram model.

**Proposition.** The merge operation can be applied in any two sequences under the following four conditions, if a Markov word *m*-gram model is adopted.

(1) The sequence length should be longer than *m*-1.
(2) The lengths of these two sequences should be equal.
(3) The last *m*-1 words in these two sequences should be equal.
(4) These two sequences should cover the same words.

*Proof*:

The first three are the basic conditions of a Markov *m*-gram model. In this model, the system uses the last *m*-1 words to predict the probability of the current word. Let the probabilities of two sequences $H_1$ and $H_2$ be $P(H_1)$ and $P(H_2)$, and $P(H_1)>P(H_2)$. When the next word $w_n$ ($n\geq m$-1) is read, their probabilities become

$P(H_1)*P(w_n|w_{1(n-m+1)}, ..., w_{1(n-1)})$ and
$P(H_2)*P(w_n|w_{2(n-m+1)}, ..., w_{2(n-1)})$, respectively.

If the last *m*-1 words are the same, i.e., $w_{1(n-m+1)}=w_{2(n-m+1)}$, ..., $w_{1(n-1)}=w_{2(n-1)}$, then the former is still larger than the latter. However, if the last m-1 words of these two sequences are not the same, then the former may be smaller than the latter. Thus, merging may preserve the sequence with smaller probability and may introduce erroneous results.

In fact, the first three conditions are enough for the other Markov-based applications such as phone-to-text transcription, *etc*. However, there is a problem in bag generation , if we do not obey the last condition. Consider a general case. Let the two sequences $H_1$ and $H_2$ have the following forms.

$H_1$: $w_{10}$, $w_{11}$, ..., $w_{1(n-m)}$, $w_{(n-m+1)}$, ..., $w_{(n-1)}$
$H_2$: $w_{20}$, $w_{21}$, ..., $w_{2(n-m)}$, $w_{(n-m+1)}$, ..., $w_{(n-1)}$

If {$w_{10}$, $w_{11}$, ..., $w_{1(n-m)}$} is not equal to {$w_{20}$, $w_{21}$, ..., $w_{2(n-m)}$}, there must exist some $w_{1i}$ and $w_{2j}$ such that $w_{1i} \neq w_{2j}$. If $P(H_1)>P(H_2)$, then the word sequence involving $w_{1i}$, i.e.,

$w_{20}$, $w_{21}$, ..., $w_{2(n-m)}$, $w_{(n-m+1)}$, ..., $w_{(n-1)}$, $w_{1i}$,

---

[1] Its time complexity is O(*n*!).

**Table 1.** Statistics of the Outside Test Data

| category | total sentences | total words | extracted sentences | extracted words |
|---|---|---|---|---|
| A | 192 | 1151 | 108 | 457 |
| G | 216 | 1166 | 156 | 685 |
| J | 323 | 1712 | 229 | 981 |
| N | 307 | 1384 | 252 | 959 |

**Table 2.** Experimental Results of Outside Test

| category | approach 1 (optimal solution) | | approach 2 (near optimal solution) | |
|---|---|---|---|---|
| | sentence correct rate | word correct rate | sentence correct rate | word correct rate |
| A | 40.74% | 49.45% | 34.26% | 38.07% |
| G | 29.49% | 44.23% | 22.44% | 32.26% |
| J | 34.50% | 43.53% | 30.13% | 37.21% |
| N | 42.86% | 49.11% | 34.13% | 38.48% |
| average correct rate | 37.18% | 46.30% | 30.47% | 36.63% |

becomes neglected. This sequence may have higher probability, so that error occurs. ∎

A newspaper corpus which includes 350775 sentences (2461178 words) is adopted as the source of the training data. It contains texts of several categories. Therefore, it can be regarded as a general corpus. The symbol * is added to the beginning and ending of all sentences. In this paper, a Markov word bigram model is considered in bag generation to generate Chinese sentences. With the Markov word bigram model, 2811953 total pairs and 905470 distinct pairs are extracted from this corpus. For outside test, four documents are selected from NTU Corpus, which is a Chinese balanced corpus. They belong to categories A (reportage), G (belles lettres), J (learned) and N (adventure). The statistics of these documents are shown in Table 1. A subset of sentences, which have 1-6 words, are selected from these documents. Columns 4 and 5 in Table 1 denote the statistics of these data. Table 2 demonstrates the experimental results of testing the extracted sentences, i.e., length 1-6.

Approach 1 uses all the four conditions in Algorithm 1, but approach 2 only uses the first three conditions. Two criteria, i.e., sentence correct rate and word correct rate, are applied to evaluating these two approaches. The former denotes how many sentences are reproduced correctly, and the latter denotes how many words occupy the correct positions. Approach 1 has better performance than approach 2 in these two aspects. However, approach 2 is more efficient than approach 1. The average performance of these two approaches is not good enough because of the small training corpus.

The document of category A is selected from newspapers, so that the performance of processing this document is better than that of categories G and J.

## 3. The Corrective Training Algorithm

In corrective training, two major issues should be considered: (1) Which object should be adjusted? and (2) How many probabilities will be reassigned to the object? These problems depend on language models and applications. This paper focuses on bag generation with Markov word bigram model. The permutation ρ is defined by formula(1). Formula (2) is derived further from formula (1).

$$\rho = \arg\max_{\rho} P(w_{\rho(0)}) * \prod_{i=0}^{n} P(w_{\rho(i+1)}|w_{\rho(i)}) \quad \ldots \ldots \ldots (1)$$

$$= \arg\max_{\rho} \frac{P(*, w_{\rho(1)}) * P(w_{\rho(1)}, w_{\rho(2)}) * \ldots * P(w_{\rho(n)}, *)}{P(w_{\rho(1)}) * P(w_{\rho(2)}) * \ldots * P(w_{\rho(n)})}$$

$$= \arg\max_{\rho} P(*, w_{\rho(1)}) * P(w_{\rho(1)}, w_{\rho(2)}) * \ldots * P(w_{\rho(n)}, *) \quad \ldots (2)$$

The denominators of all the permutations are equal, so that they can be neglected and only the probabilities of adjacent words are used instead of the original conditional probabilities. Consider two word strings D = <*, $w_1$, $w_2$, ..., $w_n$, *> and C = <*, $w_{\rho(1)}$, $w_{\rho(2)}$, ..., $w_{\rho(n)}$, *>. They correspond to the desired result and the final computed result, respectively. If C is the same as D, then no adjustment is required. Otherwise, Algorithm 1 finds the word pairs that may have to be adjusted.

```
Algorithm 1. FindPair(<*, w₁, w₂, ..., wₙ, *>,<*, wρ(1), wρ(2), ..., wρ(n), *>)
    for i = 0 to n do mark(wρ(i))=false
    for i = 0 to n do
    begin
        found=false
        j=0
        while ((j ≤ n) and (not found)) do
        begin
            if ((wᵢ = wρ(j)) and (not mark(wρ(j)))) then
            begin
                found=true
                mark(wρ(j))=true
                if (w(i+1) ≠ wρ(i+1)) then Adjust(<wᵢ,w(i+1)>,<wρ(j),wρ(j+1)>)
            end
            else j = j + 1
        end
    end
```

```
Algorithm 2. Adjust(OrderedPair,DisorderedPair)
    ΔP = P_OrderedPair - P_DisorderedPair
    if (ΔP = 0) then
    begin
        NewP_OrderedPair = α * (P_OrderedPair + Floor Value)
        NewP_DisorderedPair = α * (P_DisorderedPair - Floor Value)
        ΔP_OrderedPair = NewP_OrderedPair - P_OrderedPair
        ΔP_DisorderedPair = NewP_DisorderedPair - P_DisorderedPair
    end
    else if (ΔP < 0) then
    begin
        NewP_OrderedPair = α * (P_OrderedPair - β₁ * ΔP)
        NewP_DisorderedPair = α * (P_DisorderedPair + β₂ * ΔP)
        ΔP_OrderedPair = NewP_OrderedPair - P_OrderedPair
        ΔP_DisorderedPair = NewP_DisorderedPair - P_DisorderedPair
    end
```

Assume $D = <*, w_1, w_2, w_3, w_4, w_5, *>$ and $C = <*, w_3, w_4, w_5, w_1, w_2, *>$. Three suspicious tuples, i.e., $(<*,w_1>,<*,w_3>)$, $(<w_2,w_3>, <w_2,*>)$ and $(<w_5,*>, <w_5,w_1>)$, are identified by Algorithm 1. Because the same word may be used more than one time in the bag, the *mark* flags guarantee that the same pair cannot appear in more than one tuple. The first pair in each suspicious tuple is called the *ordered pair*, and the second pair is called the *disordered pair*. By formula (2), if the probability of the ordered pair in each suspicious tuple is larger than that of the disordered pair, then the computed result would be the desired result.

Algorithm 2 adjusts those pairs whose probabilities do not satisfy the above condition. Two adjustments, i.e., $\Delta P_{OrderedPair}$ and $\Delta P_{DisorderedPair}$, are computed to add some probabilities to the ordered pairs and subtract some probabilities from the disordered pairs. By this way, the desired result will have higher probabilities than the final computed error result.

```
Algorithm 3. CorrectiveTraining(S_1, S_2, S_3, ..., S_m)
    while not (one of the stopping criteria is met) do
    begin
        for i = 1 to m do
        begin
            Let S_i be <*, w_1, w_2, ..., w_n, *>
            ρ = formula1(<*, w_1, w_2, ..., w_n, *>)
            if <*, w_ρ(1), w_ρ(2), ..., w_ρ(n), *> ≠ <*, w_1, w_2, ..., w_n, *>
                then FindPair(<*, w_1, w_2, ..., w_n, *>,<*, w_ρ(1), w_ρ(2), ..., w_ρ(n), *>)
        end
        for each (ordered or disordered) pair do
        begin
            Compute the Average Adjustments ΔP_Pair of This Pair
            if P_Pair + ΔP_Pair < 0 then P_Pair = 0.00001 else P_Pair = P_Pair + ΔP_Pair
        end
    end
```

In Algorithm 2, $\alpha$ is the *scaling factor*. It is often set to 1. $\beta_1$ and $\beta_2$ ($0 \leq \beta_1, \beta_2 \leq 1$) are the *learning rates* of the ordered pairs and the disordered pairs, respectively. The sum of $\beta_1$ and $\beta_2$ must be greater than 1. In general, $\beta_1$ and $\beta_2$ are used to control the distance between the ordered pair and the disordered pair after adjustment. The *distance* $\delta$ is equal to $(\beta_1+\beta_2-1)*abs(\Delta P)$. The function $abs(\Delta P)$ computes the absolute value of $\Delta P$. Assume an ordered pair OP and a disordered pair DP have probabilities 0.1 and 0.6, respectively. $\beta_1$ and $\beta_2$ are set to 0.6 and 0.5, respectively. After adjustment, the probabilities of OP and DP become 0.4 and 0.35, respectively. Their difference $\delta$ is 0.05. $\delta$ is an important factor for a robust language model, and it highly depends on $\beta_1$ and $\beta_2$. Obviously, if $\beta_1$ and $\beta_2$ are all set to 0, then no feedback information is used in this algorithm[2]. On the contrary, if $\beta_1$ and $\beta_2$ are all set to 1, then the probabilities of the ordered pair and the disordered pair are exchanged mutually. $\beta_1 > \beta_2$ ($\beta_1 < \beta_2$) means Algorithm 2 emphasizes on the positive (negative) feedback information.

Algorithm 3 shows a complete corrective training algorithm for bag generation. M sentences, $S_1, S_2, S_3, ..., S_m$, are used for corrective training. This algorithm checks whether the computed result by using formula (1) is correct or not. If it is not, this algorithm adjusts the probabilities of the ordered pairs and the disordered pairs. Because a pair may be adjusted more than one time, the average of all its adjustments is computed to avoid the *overtune* problem. For example, if there are two adjustments A1 and A2 for the same pair, then the final adjustment for this pair will be $\frac{A_1 + A_2}{2}$. Moreover, if the sum of the original and the adjustment probability is less than zero, then the probability of this pair will be set to a very small value, i.e., 0.00001. Besides, a new pair with negative or zero probability will not be allowed to add into the training table. These average adjustments are fed into the old training table, and a new training table is formed. This algorithm is repeated until one of the stopping criteria is met. In practice, several criteria can be considered.

Firstly, it is based on the magnitude of *gradient of error* (GE) shown as follows:

$$GE = \sum_{i=1}^{m} \gamma_i$$

where $\gamma_i = \sum_{j=1}^{k} abs(\text{adjustment in ordered pair}_j) + abs(\text{adjustment in disordered pair}_j)$ if $S_i$ has $k$ suspicious tuples. Otherwise, $\gamma_i = 0$.

GE specifies whether the learning direction is correct or not. Clearly, if no adjustment is performed, then GE is equal to zero. Therefore, Algorithm 3 is terminated when the magnitude of gradient of error is sufficiently small. Secondly, the algorithm stops as soon as all the tests are correct, i.e., no suspicious tuples are generated. Thirdly, the algorithm stops when no feedback information is obtained, i.e., no adjustments in all the pairs. However, it does not mean the performance achieves 100%. This is because some errors are caused by the power of language model.

---
[2] The floor value is ignored in current discussions.

**Table 3.** Experimental Results by Using Approach 1 and the Corrective Training Algorithm

| stage | performance of part 1 | | performance of part 2 | | performance of part 3 | | average correct rate | |
|---|---|---|---|---|---|---|---|---|
| | sentence level | word level | sentence level | word level | sentence level | word level | sentence level | word level |
| 0 | 40.26% | 46.88% | 39.47% | 45.93% | 23.68% | 37.98% | 34.50% | 43.53% |
| 1 | 100.0% | 100.0% | 44.74% | 49.51% | 30.26% | 41.54% | 58.52% | 64.12% |
| 2 | 93.51% | 94.36% | 100.0% | 100.0% | 32.89% | 43.92% | 75.55% | 78.80% |
| 3 | 90.91% | 91.39% | 98.68% | 98.37% | 100.0% | 100.0% | 96.51% | 96.53% |

**Table 4.** Experimental Results by Using Approach 2 and the Corrective Training Algorithm

| stage | performance of part 1 | | performance of part 2 | | performance of part 3 | | average correct rate | |
|---|---|---|---|---|---|---|---|---|
| | sentence level | word level | sentence level | word level | sentence level | word level | sentence level | word level |
| 0 | 26.85% | 31.39% | 25.93% | 30.88% | 15.89% | 25.00% | 22.91% | 29.03% |
| 1 | 72.22% | 74.25% | 27.78% | 33.03% | 21.50% | 30.27% | 40.56% | 45.74% |
| 2 | 64.81% | 67.20% | 62.04% | 61.76% | 20.56% | 31.80% | 49.23% | 53.27% |
| 3 | 65.74% | 66.31% | 60.19% | 58.53% | 59.81% | 60.37% | 61.92% | 61.74% |

## 4. Experimental Results

In order to demonstrate the effect of the corrective training algorithm in different context domains, the document of category J, i.e., a technical paper, is selected in the experiment. At first, the extracted sentences of length 1-6 are partitioned into three parts (76, 76, 77). At stage 1, part one is used to do the corrective training, and parts two and three are used to test the performance. At stage 2, part two is sent to corrective training, and parts one and three test the performance. Finally, we apply the corrective training to part three, and use the other two parts to test the performance at stage 3. Table 3 shows the results by using approach 1 and the corrective training algorithm.

On the one hand, the above experiment shows this algorithm has good *generalization*. When we continue the corrective training on the subsequent part(s), the performance of the preceding part(s) remain very high. On the other hand, when the number of test sentences from the specific context domain increase, the performance of the subsequent test is improved significantly.

Next, approach 2 and the corrective training algorithm are applied to the complete document J. The three parts have 108 (567), 108 (557) and 107 (588) sentences (words), respectively. The results are shown in Table 4. Approach 2 is a near optimal bag generation so that incomplete feedback information decreases the power of the adaptive learning model. Hopefully, bag generation can be coupled with other modules in practical applications, e.g. parser in machine translation systems. Parser can partition a bag into several smaller bags (Chen and Lee, 1993b). In this way, the effectiveness is not a problem, and approach 1 (optimal bag generation) can be adopted.

## 5. Concluding Remarks

This paper proposes an corrective training algorithm for task adaptation to best-fit the run-time environment in the application of bag generation. It controls the distance of the ordered pairs and the disordered pairs in the suspicious tuples. The resulting techniques are greatly simplified and robust. They give improved performance. Although this adaptive learning algorithm is a *greedy* algorithm, i.e., *linear gradient search* algorithm, that seeks out a *local optimization* result, it still has strong probability to achieve the *global optimization* result because it starts with a good initial state, i.e., initial training table. Besides, this corrective training algorithm is also suitable for *incremental training*. Initially, training table can be generated from a generic corpus. If the test sentences come from other specific domains, this algorithm automatically revises the old training table and produces a specific training table. In general, these techniques can be easily extended to various language models and corpus-based applications.

In this paper, we assign each parameter, i.e., $\alpha$, $\beta_1$, $\beta_2$ and floor value, used in the corrective training algorithm a constant value. However, is the procedure stable for all values of these parameters, or are there universal values of these parameters? The parameter-setting problem is important and needs to further investigate. Moreover, a more robust corrective training algorithm, e.g., *non-linear corrective training algorithm*, is also demanded in the future.

## Acknowledgements

Research on this paper was partially supported by National Science Council grant NSC-83-0408-E-002-019.